\documentclass{article}
\usepackage{ijcai25}

\usepackage{times}
\usepackage{soul}
\usepackage{url}
\usepackage{multirow}
\usepackage{color}
\usepackage[hidelinks]{hyperref}
\usepackage[utf8]{inputenc}
\usepackage[small]{caption}
\usepackage{graphicx}
\usepackage{amsmath}
\usepackage{amsthm}
\usepackage{booktabs}
\usepackage{algorithm}
\usepackage{algorithmic}
\usepackage{amsmath}
\usepackage{amssymb}
\usepackage{booktabs}  % 用于 \toprule, \midrule, \bottomrule

\title{Trainability-Oriented Hybrid Quantum Regression via Geometric Preconditioning and Curriculum Optimization}

\author{
Qingyu Meng$^1$ \And
Yangshuai Wang$^2$
\affiliations
$^1$Shanghai Jiao Tong University-Chongqing Institute of Artificial Intelligence, Chongqing 401329, China.\\
$^2$Department of Mathematics, National University of Singapore, 10 Lower Kent Ridge Road, Singapore.\\
\emails
yswang@nus.edu.sg,
}
\begin{document}

\maketitle

\begin{abstract}
Quantum neural networks (QNNs) have attracted growing interest for scientific machine learning, yet in regression settings they often suffer from limited trainability under noisy gradients and ill-conditioned optimization. We propose a hybrid quantum--classical regression framework designed to mitigate these bottlenecks. Our model prepends a lightweight classical embedding that acts as a learnable geometric preconditioner, reshaping the input representation to better condition a downstream variational quantum circuit. Building on this architecture, we introduce a curriculum optimization protocol that progressively increases circuit depth and transitions from SPSA-based stochastic exploration to Adam-based gradient fine-tuning. We evaluate the approach on PDE-informed regression benchmarks and standard regression datasets under a fixed training budget in a simulator setting. Empirically, the proposed framework consistently improves over pure QNN baselines and yields more stable convergence in data-limited regimes. We further observe reduced structured errors that are visually correlated with oscillatory components on several scientific benchmarks, suggesting that geometric preconditioning combined with curriculum training is a practical approach for stabilizing quantum regression.
\end{abstract}

\section{Introduction}
\label{sec:intro}

Recent work has extended quantum neural networks (QNNs) beyond discrete classification toward regression and continuous-function approximation in scientific machine learning (SciML)~\cite{abbas2021power,schuld2014quest,beer2020training}. This interest is motivated by the expressive structure of parameterized quantum circuits and the possibility that certain circuit-induced inductive biases may be beneficial under constrained data or parameter budgets. Representative SciML tasks, including solving partial differential equations (PDEs)~\cite{xiao2024physics,farea2025qcpinn}, predicting molecular properties~\cite{lu2025quantum}, and fitting potential energy surfaces~\cite{monaco2024quantum,le2025symmetry}, often require learning physically meaningful mappings from sparse observations. In these settings, regression accuracy is typically evaluated over dense query points (e.g., collocation grids for PDEs), which makes training sensitive to optimization noise and poor conditioning. We therefore focus on data-scarce regimes where sample efficiency and optimization robustness are critical.

Despite this promise, applying QNNs to high-precision regression is often hindered by trainability issues, including barren plateaus (BP)~\cite{mcclean2018barren,cerezo2021higher,pesah2021absence} and ill-conditioned loss landscapes. The challenge is particularly pronounced in regression objectives that aggregate errors over many points (e.g., PDE collocation points), where the effective objective becomes more global and can amplify gradient noise under finite-shot estimation. This creates a practical trade-off between expressivity and trainability: increasing circuit depth can improve representational capacity, but may also exacerbate vanishing gradients and sensitivity to noise. Moreover, fixed data encodings may be suboptimal across tasks with different input geometries, potentially leading to ineffective feature extraction and unstable optimization dynamics. Empirically, these effects can manifest as high gradient variance and structured residual errors on regimes with spatially oscillatory solutions, motivating explicit strategies for improving conditioning during training.

To mitigate these issues, we propose a hybrid framework that combines a classical neural network with a variational quantum circuit to improve the stability of quantum regression. Concretely, a small multi-layer perceptron (MLP) maps inputs into a low-dimensional latent space prior to quantum encoding, acting as a learnable geometric preconditioner. By shaping the input representation, the classical module can improve conditioning for the downstream quantum ansatz and reduce the reliance on prohibitively deep circuits. To keep the classical component in a preconditioning role rather than a standalone solver, we restrict its capacity through a low-dimensional latent bottleneck and a small parameter budget. On the optimization side, we adopt a two-stage curriculum: we begin with SPSA to provide robust exploration under stochastic objectives, and then switch to Adam for gradient-based refinement once training stabilizes. This is paired with layer-wise circuit growth so that model complexity increases gradually with the optimization phase. Overall, our design is guided by a separation-of-roles hypothesis: the classical module primarily improves geometric conditioning of inputs, while the quantum circuit models residual nonlinear structure under controlled optimization dynamics.

Our main contributions are threefold:
\begin{itemize}
    \item We propose a hybrid quantum--classical regression architecture in which a lightweight classical embedding acts as a learnable geometric preconditioner. The embedding reshapes the input representation to better condition a downstream variational quantum circuit, reducing sensitivity to circuit depth and alleviating the need for overly deep feature-encoding circuits.
    \item We develop a curriculum optimization protocol that combines progressive circuit growth with an optimizer transition from SPSA to Adam. This strategy improves trainability under stochastic objectives induced by finite-shot estimation and supports stable fine-tuning of the regression objective.
    \item We present an empirical study on SciML regression benchmarks, including physics-informed solvers for multiple 2D/3D PDEs and standard regression datasets. Under a fixed training budget and simulator setting, we observe consistent improvements over pure QNN baselines and improved convergence stability alongside the considered classical references in data-limited regimes.
\end{itemize}
Our focus is on improving trainability and optimization stability for QNN regression under our experimental setting. Establishing competitiveness against stronger classical baselines and larger-scale quantum regimes is left for future work.

\section{Related Works}
\label{sec:related_works}

\textbf{Quantum Neural Networks for Continuous Domains.}
Early work in quantum machine learning (QML) emphasized discrete classification~\cite{mitarai2018quantum}, whereas more recent studies have explored QNNs for regression and continuous function approximation, which are central to SciML. Data re-uploading provides a practical construction for expressive parameterized quantum circuits and has been used to motivate function-approximation capabilities in finite-dimensional settings~\cite{perez2020data}. Building on this line, differentiable quantum circuits have been investigated for linear and nonlinear PDE solvers~\cite{cao2025quantum,kyriienko2021solving,jin2024quantum,hu2024quantum}, although training on NISQ devices is still challenged by noise, limited connectivity, and the cost of gradient estimation. Recent works have introduced trainable or adaptive embeddings to improve data encoding and empirical performance~\cite{berger2025trainable}. In parallel, theoretical studies analyze regimes where QNN-induced representations can differ from classical kernel models~\cite{jerbi2023quantum,bowles2024contextuality}. Our work complements these directions by focusing on trainability in regression: we study an architecture-and-optimization design intended to make such expressive constructions easier to train under practical budgets.

\noindent\textbf{Optimization Landscapes and Barren Plateaus.}
Barren plateaus (BP) remain a key obstacle to training deep QNNs~\cite{mcclean2018barren}, with related analyses extending to higher-order information such as ``Hessian barren plateaus''~\cite{wang2024investigating}. Recent results highlight that trainability depends on the joint effects of circuit structure, noise, and initialization~\cite{ragone2024lie}. Proposed mitigation strategies include identity-style initialization~\cite{cerezo2021cost} and overparameterization~\cite{larocca2023theory}, though the latter may increase circuit depth and resource requirements in ways that are unfavorable for near-term settings. In contrast, we emphasize improved optimization stability under a fixed training budget by combining a layer-wise curriculum~\cite{skolik2021layerwise} with a two-stage optimizer schedule, aiming to balance expressivity and trainability without relying on very deep or heavily overparameterized circuits.

\noindent\textbf{Hybrid Architectures and Geometric Preconditioning.}
Hybrid quantum--classical architectures are widely used in near-term settings, both to exploit available quantum resources and to reduce the burden of purely quantum feature learning. Prior work has investigated variational regression and quantum-enhanced solvers~\cite{wang2023variational}, as well as empirical comparisons of optimizers for QNN training~\cite{wiedmann2023empirical}. Complementary to optimizer-centric studies, emerging research suggests that adaptive embeddings can act as implicit preconditioners by improving how inputs are presented to the quantum circuit~\cite{multiple2025embedding}. Our contribution is to make this connection explicit: we tightly couple a capacity-controlled classical embedding with the quantum ansatz and pair it with curriculum optimization. In our terminology, \emph{geometric preconditioning} refers to learning an input transformation that improves the conditioning of the downstream quantum regression objective (and hence optimization stability), rather than merely increasing representational capacity.

\section{Methodology}
\label{sec:methodology}

\begin{figure*}[htbp]
    \centering
    \includegraphics[width=1.0\textwidth]{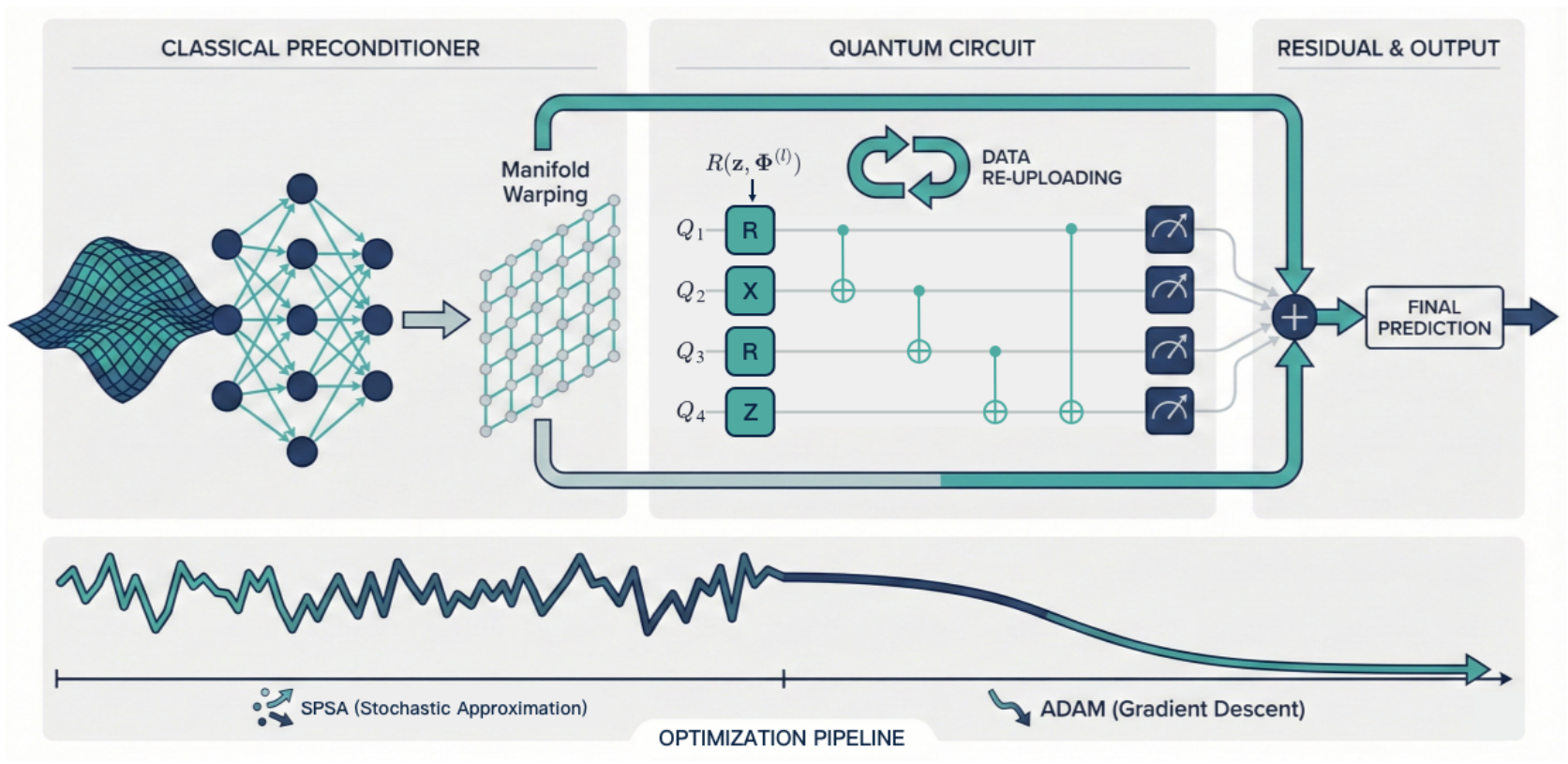}
    \caption{\textbf{Hybrid Quantum Regression Framework.} A lightweight classical embedding transforms inputs before quantum encoding to improve conditioning for the variational circuit. Training follows a curriculum that gradually increases circuit depth and transitions from SPSA-based exploration to Adam-based refinement.}
    \label{fig:schematic}
\end{figure*}

We propose a framework that targets optimization instability in quantum regression via two complementary components. First, we introduce a capacity-controlled classical embedding that acts as a \textit{geometric preconditioner}, reshaping the input representation to improve the conditioning of a downstream variational quantum circuit. Second, we adopt a \textit{curriculum-driven} training protocol that schedules model capacity and optimization dynamics during learning. The overall workflow is illustrated in Figure~\ref{fig:schematic}.

\subsection{Hybrid Architecture with Classical Embedding}
\label{sec:sub:hybrid_arch}

Encoding high-dimensional inputs directly into quantum states can require deep feature-encoding circuits, which may aggravate trainability challenges such as barren plateaus under finite-shot noise and hardware imperfections~\cite{mcclean2018barren}. To mitigate this, we prepend a classical embedding network $f_{\theta_c}: \mathbb{R}^d \to \mathbb{R}^p$, implemented as a lightweight MLP:
\begin{equation}
    \mathbf{z} = f_{\theta_c}(\mathbf{x}).
\end{equation}
We choose the latent dimension $p$ to match the input dimensionality required by the quantum feature map (typically $p=n_q$ for single-qubit encoding gates). We view this mapping as a \textit{learnable geometric preconditioner}: it learns an input transformation that can improve the conditioning of the quantum regression objective with respect to the circuit parameters. Intuitively, optimizing $\theta_c$ reshapes the representation geometry presented to the variational ansatz, reducing sensitivity to circuit depth and easing optimization compared to fixed encodings.

To keep the classical module in a preconditioning role rather than a standalone solver, we explicitly restrict its capacity using a low-dimensional latent bottleneck and a small parameter budget, and we evaluate its contribution via ablations in Section~\ref{sec:experiments}. This design encourages a complementary interaction: the embedding captures coarse structure in the input--output mapping, while the quantum circuit provides a flexible nonlinear component that can refine the prediction when direct optimization of a deeper circuit is difficult.

The latent vector $\mathbf{z}$ is then encoded by a parameterized circuit $U_{\theta_q}(\mathbf{z})$ (defined in~\eqref{eq:Uq}) acting on $n_q$ qubits. We obtain a scalar quantum feature by measuring a local observable $\hat{O}$ (e.g., Pauli-$Z$ on the first qubit):
\begin{equation}
\begin{split}
    &y_q = \langle \psi(\mathbf{z}, \theta_q) | \hat{O} | \psi(\mathbf{z}, \theta_q) \rangle, \\
    &|\psi(\mathbf{z}, \theta_q) \rangle = U_{\theta_q}(\mathbf{z}) | 0 \rangle^{\otimes n_q}.
\end{split}
\end{equation}
To form the final prediction $\hat{y}$, we use a residual-style readout: we concatenate the classical latent features $\mathbf{z}$ with the quantum feature $y_q$ and apply a linear readout,
\begin{equation}
    \hat{y} = \mathbf{w}^\top [\mathbf{z}, y_q] + b.
\end{equation}
This choice preserves a simple global trend model through the classical pathway while using the quantum circuit as a compact nonlinear correction term.

\subsection{Trainable Quantum Feature Map via Data Re-uploading}
\label{sec:sub:encoding}

To encode the latent representation $\mathbf{z}$ into the quantum Hilbert space, we employ a data re-uploading strategy with trainable scaling parameters~\cite{perez2020data}. Compared with fixed encoding schemes (e.g., standard angle encoding with static coefficients), data re-uploading introduces additional trainable degrees of freedom in the encoding map, which can increase expressivity and provide flexibility for optimization under a fixed circuit template.

We construct $U_{\theta_q}(\mathbf{z})$ as a sequence of $L$ layers. Each layer $\ell$ consists of a feature-dependent rotation block $U_{\mathrm{rot}}^{(\ell)}$ followed by a fixed entangling block $U_{\mathrm{ent}}^{(\ell)}$. The overall unitary is
\begin{equation}\label{eq:Uq}
    U_{\theta_q}(\mathbf{z}) = \prod_{\ell=L}^{1} \left( U_{\mathrm{ent}}^{(\ell)} \cdot U_{\mathrm{rot}}^{(\ell)}(\mathbf{z}, \phi^{(\ell)}, \beta^{(\ell)}) \right),
\end{equation}
where the product index $\ell=L \to 1$ follows standard operator ordering.

The rotation block applies single-qubit rotations parameterized by the latent inputs $\mathbf{z}$ and trainable parameters $\phi^{(\ell)}$ and $\beta^{(\ell)}$:
\begin{equation}
    U_{\mathrm{rot}}^{(\ell)}(\mathbf{z}, \phi^{(\ell)}, \beta^{(\ell)}) = \bigotimes_{j=1}^{n_q} R\!\left(\phi^{(\ell)}_{j} z_j + \beta^{(\ell)}_j\right).
\end{equation}
Here, $R(\cdot)$ denotes a single-qubit rotation gate; in our implementation, we default to $R_Y(\cdot)$ unless otherwise specified. The entangling block $U_{\mathrm{ent}}^{(\ell)}$ uses a fixed CNOT topology (e.g., linear or circular) to introduce correlations across qubits. We denote the full set of trainable circuit parameters by $\theta_q = \{ \phi^{(\ell)}, \beta^{(\ell)} \}_{\ell=1}^L$.

Connections between data re-uploading and spectral representations have been studied in prior work; for example, QNNs with re-uploading can be related to truncated Fourier-like expansions under suitable assumptions~\cite{yu2022power}. From this perspective, the scaling factors $\phi^{(\ell)}$ act as learnable frequency controls, while $\beta^{(\ell)}$ adjusts phases and offsets. Intuitively, optimizing $\phi$ can stretch or compress the effective input coordinates seen by the circuit, which may help match the frequency content required by the target mapping. This motivation is particularly relevant in PDE-informed regression, where solutions can exhibit oscillatory behavior; we empirically evaluate this setting in Section~\ref{sec:experiments}.

\subsection{Curriculum-Driven Optimization Protocol}
\label{sec:sub:dynamics}

\paragraph{Training objective.}
Let $\theta = \theta_c \cup \theta_q$ denote the complete set of trainable parameters, combining the classical weights $\theta_c$ and the quantum variational parameters $\theta_q$. We minimize a task-dependent objective $\mathcal{L}(\theta)$ that measures approximation error.
For supervised regression with dataset $\{(\mathbf{x}_i,y_i)\}_{i=1}^N$, we use the mean squared error
\begin{equation}
\label{eq:loss_mse}
\mathcal{L}(\theta)
=
\frac{1}{2N}\sum_{i=1}^{N}\big(\hat y(\mathbf{x}_i;\theta)-y_i\big)^2.
\end{equation}

For PDE tasks, we minimize a mean-squared residual over interior collocation points $\{\mathbf{x}_m\}_{m=1}^{N_r}$,
\begin{equation}
\label{eq:loss_pde}
\mathcal{L}_{\mathrm{res}}(\theta)
=
\frac{1}{2N_r}\sum_{m=1}^{N_r}
\left\|
\mathcal{N}[\hat u_\theta](\mathbf{x}_m)-f(\mathbf{x}_m)
\right\|_2^2,
\end{equation}
where $\mathcal{N}$ denotes the differential operator, $f$ is the forcing term, and $\hat u_\theta$ is the learned solution surrogate. In our experiments, Dirichlet boundary conditions are incorporated via the hard-constraint ansatz described in Section~\ref{sec:sub:setup}, so $\mathcal{L}_{\mathrm{res}}$ is optimized over interior points.

\paragraph{Curriculum optimization.}
The resulting objective is nonconvex due to the hybrid classical--quantum parametrization, and its evaluation is stochastic under minibatching and finite-shot quantum measurements. Moreover, variational quantum circuits can exhibit vanishing-gradient regimes for certain architectures and depth scalings. Motivated by these trainability challenges, we adopt a two-stage protocol that transitions from stochastic exploration to gradient-based refinement, coupled with a progressive increase in circuit depth. For each depth configuration, we first run SPSA for $T_{\mathrm{SPSA}}$ iterations to improve robustness to noisy objective evaluations, and then switch to Adam for fine-tuning until a convergence criterion on the training loss is met. The procedure is summarized in Algorithm~\ref{alg:hybrid_summary}.

\begin{algorithm}[H]
\caption{Hybrid Curriculum Optimization}
\label{alg:hybrid_summary}
\begin{algorithmic}[1]
\STATE \textbf{Input:} Initial depth $L=1$, maximum depth $L_{\max}$, parameters $\theta$.
\WHILE{$L \le L_{\max}$}
    \STATE \textbf{Stage 1 (SPSA):} Update $\theta$ using SPSA for $T_{\mathrm{SPSA}}$ iterations.
    \STATE \textbf{Stage 2 (Adam):} Fine-tune $\theta$ using hybrid gradients until convergence.
    \IF{$L < L_{\max}$}
        \STATE \textbf{Growth:} Add layer $L+1$ to the quantum circuit.
        \STATE \textbf{Init:} Initialize new parameters $\theta_{\mathrm{new}} \approx 0$ (identity-style initialization).
        \STATE $\theta \leftarrow \theta \cup \theta_{\mathrm{new}}$, \ $L \leftarrow L + 1$.
    \ENDIF
\ENDWHILE
\end{algorithmic}
\end{algorithm}

\paragraph{Stage 1: Stochastic approximation via SPSA.}
We initiate the training of each circuit configuration using SPSA. In contrast to analytic gradient methods based on the parameter-shift rule, which typically require two circuit evaluations per quantum parameter (and per measured observable), in addition to minibatch aggregation, SPSA estimates a descent direction using only two objective evaluations per iteration, independent of the parameter dimension. This property can reduce per-iteration circuit calls and can be advantageous when objective evaluations are noisy due to finite shots and stochastic minibatches. At iteration $t$, the SPSA gradient estimator is
\begin{equation}
\label{eq:spsa}
\hat{\mathbf{g}}_t(\theta_t)
=
\frac{\mathcal{L}(\theta_t + c_t \Delta_t) - \mathcal{L}(\theta_t - c_t \Delta_t)}{2c_t}\;\Delta_t^{-1},
\end{equation}
where $\Delta_t \in \{-1, 1\}^{d_\theta}$ is a perturbation vector with i.i.d.\ Rademacher entries and $\Delta_t^{-1}$ denotes elementwise inversion.
For Rademacher perturbations, $\Delta_{t,i}^{-1}=\Delta_{t,i}$ for all $i$, so \eqref{eq:spsa} matches the standard SPSA update rule. Under standard smoothness assumptions, $\mathbb{E}[\hat{\mathbf{g}}_t(\theta_t)]$ approximates $\nabla \mathcal{L}(\theta_t)$ up to $\mathcal{O}(c_t^2)$ bias while requiring only two objective evaluations per iteration. In our experiments, this stochastic phase improves optimization stability before fine-tuning.

\paragraph{Stage 2: Analytic fine-tuning with identity initialization.}
After the SPSA phase, we switch to the Adam optimizer. Gradients are computed via a hybrid differentiation strategy: the classical parameters $\theta_c$ are updated using standard backpropagation, while the quantum parameters $\theta_q$ use analytic parameter-shift differentiation~\cite{wierichs2022general}.

Let $y_q(\theta_q)$ denote the quantum measurement output (an expectation value) used by the hybrid model. For a quantum parameter $\theta_j \in \theta_q$ that enters a gate of the form $\exp(-i\theta_j P/2)$ with $P^2=I$ (and more generally via the generalized shift rule~\cite{wierichs2022general}), the parameter-shift rule yields
\begin{equation}
\label{eq:ps}
\frac{\partial y_q}{\partial \theta_j}
=
\frac{1}{2}\left[
y_q\!\left(\theta_q+\frac{\pi}{2}\mathbf e_j\right)
-
y_q\!\left(\theta_q-\frac{\pi}{2}\mathbf e_j\right)
\right].
\end{equation}
To connect \eqref{eq:ps} with the loss, we use the linear readout $\hat{y} = \mathbf{w}^\top [\mathbf{z}, y_q] + b$. Let $w_q$ denote the scalar component of $\mathbf w$ multiplying $y_q$, so that $\partial \hat y/\partial y_q = w_q$. Then the chain rule gives
\begin{equation}
\label{eq:chain}
\frac{\partial \mathcal{L}}{\partial \theta_j}
=
\frac{\partial \mathcal{L}}{\partial \hat y}\;
w_q\;
\frac{\partial y_q}{\partial \theta_j}.
\end{equation}
Analytic fine-tuning typically requires two circuit evaluations per quantum parameter per measured observable (and per minibatch aggregation), which motivates using SPSA during early exploration. With finite shots, the shifted expectation values in \eqref{eq:ps} are estimated stochastically, yielding a gradient estimate whose variance depends on the measurement budget.

\paragraph{Layer-wise growth strategy.}
We employ a \textit{layer-wise growth strategy}. When the circuit depth increases from $L$ to $L+1$, parameters of the newly added layer are initialized near zero. For parameterizations where zero angles correspond to the identity (e.g., $R_Y(0)=I$), the added layer $U_{\mathrm{new}}(\theta_{\mathrm{new}})$ satisfies $U_{\mathrm{new}}(\theta_{\mathrm{new}})\approx I$ for small $\|\theta_{\mathrm{new}}\|$. This continuity reduces the risk of disrupting previously learned solutions and helps maintain stable optimization during depth growth.

\section{Numerical Experiments}
\label{sec:experiments}

We evaluate the proposed framework on two categories of regression problems: (1) PDE-informed regression under the Physics-Informed Neural Network (PINN) paradigm, and (2) standard tabular regression benchmarks with limited data. Our evaluation focuses on predictive accuracy and convergence stability. All numerical simulations presented in this work were implemented using the MindSpore Quantum framework~\cite{xu2024mindspore}, which provides a powerful environment for quantum computation simulation. The source code and data required to reproduce the numerical results reported in this paper are publicly available at \url{https://github.com/YangshuaiWang/QNN-Regression-Mindspore.git}.

\subsection{Experimental Setup}
\label{sec:sub:setup}

\paragraph{Scientific Tasks.}
We consider four PDE benchmarks on $\Omega = [-1, 1]^d$ with homogeneous Dirichlet boundary conditions. To satisfy the boundary conditions exactly, we use a hard-constraint ansatz
$\hat{u}(\mathbf{x}) = \mathcal{N}_{\theta}(\mathbf{x}) \prod_{i=1}^{d} (1 - x_i^2)$,
where $\mathcal{N}_{\theta}$ denotes the raw network output. The benchmarks are selected to probe different sources of difficulty (dimension, nonlinearity, and oscillatory structure):

\begin{itemize}
    \item \textbf{2D Poisson:} $-\Delta u = f$ with $u(\mathbf{x}) = \sin(\pi x)\sin(\pi y)$.
    \item \textbf{2D Nonlinear:} $-\Delta u + u^3 = f$ with $u(\mathbf{x}) = \sin(2\pi x)\sin(3\pi y)$.
    \item \textbf{2D Convection--Diffusion:} $\boldsymbol{\beta} \cdot \nabla u - \epsilon \Delta u = f$, with $\boldsymbol{\beta}=[3, 3]^\top$ and $\epsilon=0.1$, and $u(\mathbf{x}) = \sin(2\pi x)\sin(2\pi y)$.
    \item \textbf{3D Modified Helmholtz:} $-\Delta u + u = f$ with $u(\mathbf{x}) = \sin(\pi x)\sin(\pi y)\sin(\pi z)$.
\end{itemize}

Models are trained by minimizing a mean-squared PDE residual over interior collocation points, using the above hard constraint to handle boundary conditions. To emulate data-limited regimes (e.g., sparse sensing), we restrict the collocation resolution to $N_r \approx 30^2$ for 2D tasks and $N_r = 12^3$ for the 3D task. For PDE training, we use Adam with a standard learning-rate schedule (coarse phase $\eta \approx 10^{-2}$ followed by fine-tuning $\eta \approx 10^{-3}$); the proposed curriculum protocol (SPSA$\rightarrow$Adam with depth growth) is applied as described in Section~\ref{sec:sub:dynamics}. In addition to reporting the training objective, we evaluate solution quality using the relative $L^2$ error against the analytical ground truth:
\begin{equation}
    \text{Error}_{L^2} = \frac{\| \hat{u} - u \|_2}{\| u \|_2}.
\end{equation}

\noindent\textbf{Regression Benchmarks and Baselines.}
For tabular regression, we use three datasets from the UCI repository~\cite{asuncion2007uci}: \textit{Yacht Hydrodynamics} ($N=308, d=6$), \textit{Energy Efficiency} ($N=768, d=8$), and \textit{Concrete Strength} ($N=1030, d=8$). We standardize inputs via z-score normalization and normalize targets to the unit interval during training; RMSE is reported on the original scale after inverse transformation. We use 5-fold cross-validation and report mean$\pm$std across folds. Unless otherwise stated, we follow a fixed training protocol without extensive per-dataset hyperparameter tuning.

We compare the proposed hybrid model against:
\begin{enumerate}
    \item \textbf{Pure QNN:} the same variational ansatz and data re-uploading strategy, but without the classical preconditioning stage (raw inputs are encoded directly by rotation gates).
    \item \textbf{Classical references:} a standard MLP, support vector regression (SVR), and random forests (RF). For PDE tasks, the MLP serves as a classical PINN reference.
\end{enumerate}
For differentiable models (Hybrid, Pure QNN, and MLP), we match parameter budgets as closely as possible; we note that this does not necessarily equate compute budgets, especially for quantum gradients under finite-shot estimation. For quantum architectures, we use $n_q \in \{4, 6\}$ qubits with depths $L \in \{2, 4\}$. The MLP baseline and the preconditioning module use two hidden layers (typically 32 and 16 units) with $\tanh$ activations.

All experiments are repeated with multiple random seeds; we report mean$\pm$std and provide implementation details (including shot budgets and optimizer hyperparameters) in the Supplementary Material.

\subsection{Efficiency Metrics}
\label{sec:sub:metrics}

To summarize accuracy--cost trade-offs under our experimental setup, we report two \emph{descriptive} composite metrics. We emphasize that these scores are intended for \emph{relative comparisons within the same implementation and simulator setting} and should not be interpreted as platform-independent measures of efficiency or as evidence of quantum advantage.

\noindent\textbf{Resource-Adjusted Error (RAE).}
RAE combines prediction error with coarse resource proxies and is used as a compact summary statistic:
\begin{equation}
    \text{RAE} = \text{RMSE} \times \left( \frac{\tilde{T} \cdot S \cdot G}{C_0} \right)^\alpha,
\end{equation}
where $\tilde{T}$ denotes wall-clock time measured under a fixed software/hardware environment, $S$ is the total number of measurement shots, and $G$ is the per-execution gate count of the quantum circuit.\footnote{Wall-clock time is sensitive to implementation details and hardware; we therefore treat $\tilde{T}$ as a descriptive quantity and also report $S$ and $G$ separately.}
The normalization constant $C_0$ is set to a reference cost (e.g., the MLP baseline under the same environment) to obtain unitless scaling, and $\alpha$ (default $0.5$) controls the strength of the cost penalty. Lower RAE indicates a better error--cost balance under our setup.
For clarity, we primarily use \emph{relative RAE} to compare quantum variants, while classical baselines serve as reference points under the chosen normalization.

\noindent\textbf{Quantum Cost--Benefit Score (QCB).}
To quantify whether a quantum model yields error improvements relative to a time proxy within the same environment, we define
\begin{equation}
    \text{QCB} = \frac{\text{Error}_{\text{ref}} - \text{Error}_{\text{QNN}}}{\max\!\left(1, \tilde{T}_{\text{QNN}} / \tilde{T}_{\text{ref}}\right)}.
\end{equation}
Here, ``ref'' denotes a chosen classical reference model trained under the same protocol. QCB measures error reduction per unit of relative time increase under our experimental setting. Positive values indicate that the QNN improves the chosen error metric without a disproportionate increase in measured time, whereas values near zero or negative indicate limited utility under the current cost proxy. We stress that QCB is a pragmatic score for within-setting comparisons rather than a claim of quantum advantage.

\subsection{Numerical Performance Analysis}
\label{sec:sub:results}

\begin{figure*}[t]
    \centering
    \includegraphics[width=1.0\textwidth]{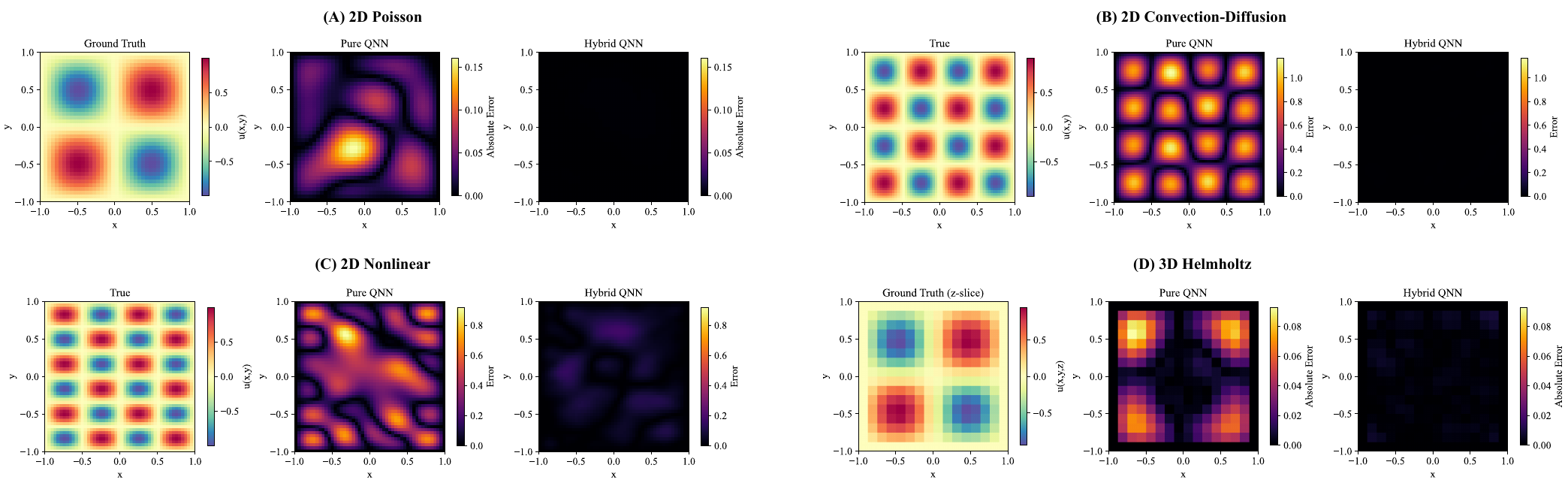}
    \caption{\textbf{Numerical validation on PDE benchmarks.}
We show ground truth and absolute error distributions for four benchmarks:
\textbf{(A)} 2D Poisson,
\textbf{(B)} 2D Convection--Diffusion,
\textbf{(C)} 2D Nonlinear Poisson, and
\textbf{(D)} 3D Modified Helmholtz. Under the fixed training protocol, the Hybrid QNN yields lower and less structured error patterns compared to the considered baselines.}
    \label{fig:pde_errors}
\end{figure*}

We evaluate the proposed framework across four PDE benchmarks (Figure~\ref{fig:pde_errors}) and three tabular regression datasets (Table~\ref{tab:uci_results}). Our goal is to assess predictive accuracy and convergence stability under the fixed experimental protocol described in Section~\ref{sec:sub:setup}. We compare the Hybrid QNN against (i) a Pure QNN that removes the classical preconditioning stage while keeping the same quantum ansatz, and (ii) classical reference models with comparable parameter budgets for differentiable baselines.

\noindent\textbf{PDE Performance and Convergence Stability.}
We report the relative $L^2$ error for all PDE benchmarks in Table~\ref{tab:pde_l2}. To reflect generalization beyond the training collocation points, all errors are computed on a dense evaluation grid ($100 \times 100$ for 2D tasks and $30^3$ for the 3D task), distinct from the training set. All methods are trained with the same collocation set and optimizer schedule (2000 epochs), and we report the model state at the final epoch to emphasize convergence behavior under a fixed training budget.

Across the four PDEs, the Hybrid QNN achieves lower relative $L^2$ error than the Pure QNN and improves upon the MLP-PINN reference under the matched protocol. The gains are most pronounced on benchmarks with more oscillatory structure (e.g., convection--diffusion and the nonlinear case), where the Pure QNN frequently exhibits slow progress early in training. While we do not perform a full frequency-resolved error decomposition, the error heatmaps in Figure~\ref{fig:pde_errors} suggest that the Hybrid model reduces structured residual patterns that are visually correlated with oscillatory solution regimes. Overall, these results are consistent with the hypothesis that preconditioning and curriculum training improve the trainability of quantum regression in PDE-informed settings. 

\begin{table}[htbp]
\centering
\caption{\textbf{PDE Solver Accuracy (Relative $L^2$ Error).}
Comparison across four benchmarks. Errors are evaluated on a dense test grid distinct from training points.}
\label{tab:pde_l2}
\resizebox{\columnwidth}{!}{
\begin{tabular}{lccc}
\toprule
Benchmark & MLP-PINN & Pure QNN & Hybrid QNN \\
\midrule
2D Poisson          & $1.5 \times 10^{-2}$ & $3.2 \times 10^{-1}$ & $\mathbf{8.5 \times 10^{-3}}$ \\
2D Convection-Diff. & $4.5 \times 10^{-2}$ & $8.1 \times 10^{-1}$ & $\mathbf{1.2 \times 10^{-2}}$ \\
2D Nonlinear        & $3.8 \times 10^{-2}$ & $7.5 \times 10^{-1}$ & $\mathbf{9.2 \times 10^{-3}}$ \\
3D Helmholtz        & $6.2 \times 10^{-2}$ & $5.4 \times 10^{-1}$ & $\mathbf{2.1 \times 10^{-2}}$ \\
\bottomrule
\end{tabular}
}
\end{table}

\noindent\textbf{Interpreting the Improvement.}
The Pure QNN (Middle Columns in Figure~\ref{fig:pde_errors}) often produces larger, structured error patterns under the same depth and training budget, suggesting sensitivity to initialization and optimization noise in these settings. Introducing the classical embedding yields a more stable optimization trajectory and improves final error. We interpret this effect as consistent with the role of the embedding as a learnable preconditioner that reshapes the input representation presented to the quantum circuit. We avoid attributing the improvement to a single factor (e.g., ``high-frequency resolution'') without frequency-resolved analysis; instead, we present the error maps and ablations as empirical evidence that the hybrid design is easier to optimize under the considered protocol.

\noindent\textbf{Generalization on Data-Limited Tabular Regression.}
Table~\ref{tab:uci_results} reports test RMSE using 5-fold cross-validation on three UCI datasets. On Yacht Hydrodynamics ($N\approx 300$), the Hybrid QNN attains $0.46 \pm 0.05$, improving over the Pure QNN ($1.31 \pm 0.42$) and outperforming the MLP baseline under our protocol ($8.08 \pm 1.12$). We note that tabular regression performance can be sensitive to baseline tuning and regularization choices. As a sanity check, we verified that the high test error of the MLP persists across folds under standard tuning within our fixed budget, despite low training RMSE, indicating overfitting in this small-data regime. The Hybrid QNN also exhibits lower variance than the Pure QNN across datasets, consistent with improved optimization stability from preconditioning.

On Energy Efficiency, the Hybrid QNN achieves the lowest RMSE ($0.36 \pm 0.04$). On Concrete Strength, random forests remain strong, which is expected for tabular data; nevertheless, the Hybrid QNN improves over differentiable baselines (MLP and SVR) under the same protocol. This distinction is relevant for settings where differentiability is required (e.g., PDE-informed objectives).

\begin{table}[htbp]
\centering
\caption{\textbf{Benchmarking Results on UCI Datasets.}
Test RMSE (mean$\pm$std) over 5-fold cross-validation.}
\label{tab:uci_results}
\resizebox{\columnwidth}{!}{
\begin{tabular}{lccccc}
\toprule
Dataset & MLP & SVR & RF & Pure QNN & Hybrid QNN \\
\midrule
Yacht    & $8.08 \pm 1.12$ & $4.83 \pm 0.55$ & $0.89 \pm 0.08$ & $1.31 \pm 0.42$ & $\mathbf{0.46 \pm 0.05}$ \\
Energy   & $3.43 \pm 0.21$ & $2.57 \pm 0.18$ & $0.57 \pm 0.04$ & $0.47 \pm 0.15$ & $\mathbf{0.36 \pm 0.04}$ \\
Concrete & $12.41 \pm 1.50$ & $6.09 \pm 0.33$ & $\mathbf{4.46 \pm 0.25}$ & $5.98 \pm 0.88$ & $5.21 \pm 0.18$ \\
\bottomrule
\end{tabular}
}
\end{table}

\noindent\textbf{Ablation Studies: Optimization and Preconditioning.}
We ablate both the optimization protocol and the classical preconditioning module. On Yacht, using Adam alone yields faster early progress but often converges to suboptimal solutions under our budget (RMSE $0.58 \pm 0.12$), while SPSA alone is more robust but converges slowly (RMSE $0.89 \pm 0.09$). The proposed two-stage protocol achieves the best performance under the same stopping criterion (RMSE $0.46 \pm 0.05$). Removing the classical embedding (Pure QNN) increases error from $0.46$ to $1.31$, supporting the role of the embedding in improving optimization stability and representation conditioning for the downstream quantum circuit.

\subsection{Resource Efficiency Analysis}
\label{sec:sub:pareto}

We conclude by summarizing the accuracy--cost trade-off under our simulator and implementation setting. Since wall-clock time is platform-dependent, our goal is not to claim platform-independent efficiency, but to provide descriptive within-setting comparisons using the composite metrics introduced in Section~\ref{sec:sub:metrics} (RAE and QCB), together with the underlying resource proxies (e.g., circuit depth, gate counts, and shot budgets).

Table~\ref{tab:rae} reports these metrics on the Concrete dataset, using the cross-validated RMSE values from Table~\ref{tab:uci_results}. For clarity, we report a relative RAE normalized by the MLP baseline under the same environment. Under this protocol, the Pure QNN incurs additional quantum overhead (e.g., parameter-shift evaluations and deeper feature encoding) while achieving limited further gains over classical references. In contrast, the Hybrid QNN achieves a lower RMSE than the Pure QNN and a lower relative RAE, indicating a more favorable error--cost balance under our cost proxy. The QCB score is positive for the Hybrid QNN with respect to the chosen reference, consistent with improved error reduction per unit of relative time increase in this setting. We stress that these summaries are descriptive and should be interpreted within the same experimental environment.

\begin{table}[t]
\centering
\caption{\textbf{Efficiency Analysis (Concrete Dataset).}
Computational trade-off summaries based on 5-fold CV RMSE. We report relative RAE normalized by the MLP baseline and QCB defined in Section~\ref{sec:sub:metrics} under the same simulator/implementation setting.}
\label{tab:rae}
\begin{small}
\begin{tabular}{lccc}
\toprule
Method & RMSE (Mean) & Relative RAE & QCB \\
\midrule
MLP (Baseline) & 12.41 & 1.00 & -- \\
Pure QNN & 5.98 & 1.15 & -0.05 \\
Hybrid QNN & \textbf{5.21} & \textbf{0.78} & \textbf{+0.85} \\
\bottomrule
\end{tabular}
\end{small}
\end{table}

\section{Conclusion and Future Work}
\label{sec:conclusion}

We presented a trainability-oriented framework for quantum regression that targets optimization instability associated with barren plateaus and ill-conditioned objectives. Our approach combines a capacity-controlled classical embedding, interpreted as a learnable geometric preconditioner, with a curriculum-driven training protocol that schedules circuit depth and optimizers during learning. Empirically, under our simulator and fixed-budget setting, this hybrid design improves over pure QNN baselines across the considered PDE-informed and tabular regression tasks and exhibits more stable convergence in data-limited regimes. Rather than relying on depth-centric overparameterization, our results suggest that jointly designing representations and optimization schedules can be a practical direction for stabilizing quantum regression in near-term settings.

\paragraph{Future work.}
First, an important next step is to evaluate the framework on NISQ hardware and study how hardware noise and error-mitigation strategies affect the observed trainability gains~\cite{temme2017error}. Second, we plan to extend the methodology to operator learning settings to test whether the proposed preconditioning-and-curriculum recipe transfers to resolution-invariant mappings~\cite{kovachki2023neural,wangquanonet}. Finally, developing theoretical characterizations of how the classical preconditioner influences optimization geometry, for example through the spectrum of the quantum Fisher information matrix (QFIM) and its relation to gradient variance, could provide deeper insight into when and why hybrid preconditioning improves trainability~\cite{liu2020quantum}. We will further strengthen validation by (i) adding stronger classical baselines (e.g., tuned gradient-boosted decision trees and Fourier-feature PINNs) under matched budgets, and (ii) performing frequency-resolved error analyses together with platform-agnostic resource accounting (circuit calls and shot budgets) and scaling studies in qubit/depth.

\appendix

\section*{Acknowledgments}
\label{sec:ack}
This research is supported by the CPS-Yangtze Delta Region Industrial Innovation Center of Quantum and Information Technology through the MindSpore Quantum Open Fund. We gratefully acknowledge the technical support from the MindSpore Quantum team and valuable discussions with collaborators during the development of this work.

\bibliographystyle{named}
\bibliography{ijcai25}

\clearpage

\section{Supplementary Numerical Results}
\label{sec:appendix_results}

This supplementary material provides additional analyses of optimization dynamics and geometry-aware updates under our experimental setting. We visualize representative training trajectories to motivate the two-stage curriculum, report aggregated statistics across random seeds, and provide sanity checks for baseline behavior. We emphasize that these analyses are descriptive and intended to support the empirical observations in the main text rather than to serve as formal guarantees.

\subsection{Detailed Analysis of Optimization Dynamics}
\label{app:optimization_ablation}

\begin{figure}[htbp]
    \centering
    \includegraphics[width=0.45\textwidth]{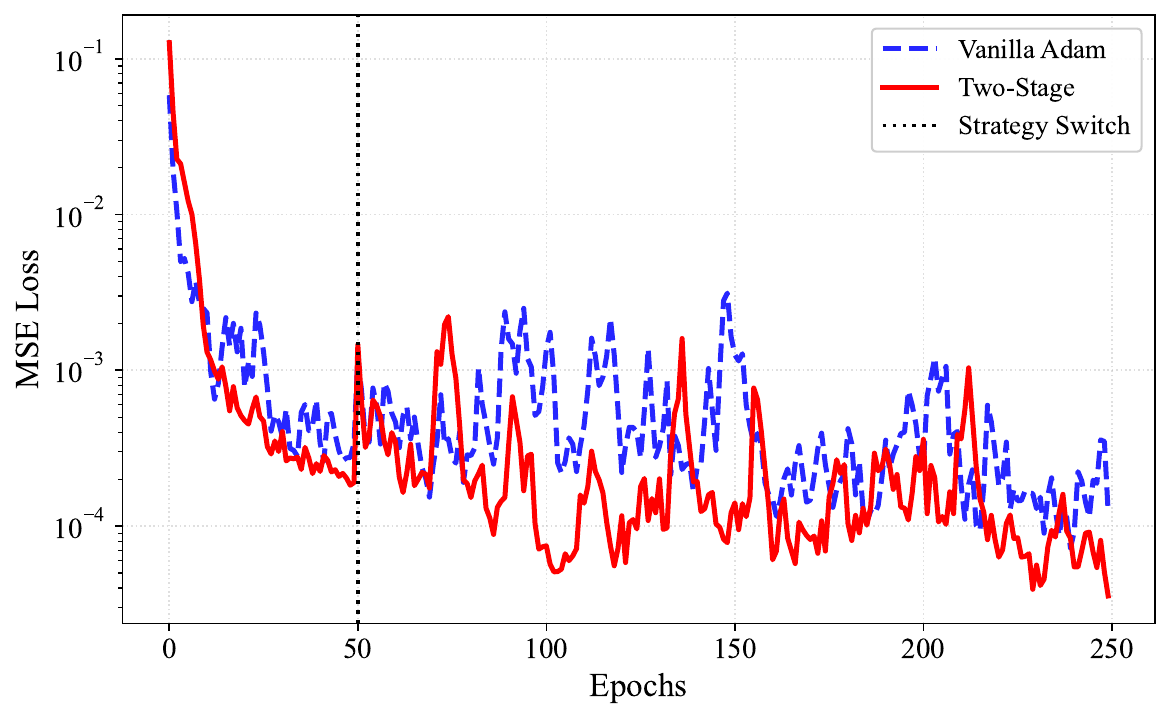}
    \caption{\textbf{Optimization dynamics on Yacht (representative fold).}
Training loss trajectories for SPSA-only, Adam-only, and the proposed two-stage schedule.}
    \label{fig:ablation}
\end{figure}

Optimization stability in hybrid quantum regression can be sensitive to initialization and stochasticity (minibatching and finite-shot estimation). To study the effect of our curriculum-driven protocol, we perform an ablation that isolates the roles of the stochastic exploration phase and the gradient-based refinement phase. Figure~\ref{fig:ablation} shows the loss evolution for one representative fold of the Yacht Hydrodynamics dataset, and Table~\ref{tab:ablation} reports summary statistics averaged over 5 random seeds. While Figure~\ref{fig:ablation} presents a representative fold, we observed qualitatively similar training-curve patterns across folds and seeds.

The Adam-only baseline (blue dashed curve) typically shows rapid initial progress but often plateaus at a suboptimal level under our fixed training budget, consistent with sensitivity to local landscape structure. In contrast, SPSA-only training is more robust to noisy objective evaluations but converges more slowly and exhibits higher variance across runs. The proposed two-stage strategy combines these behaviors: SPSA provides a noisy but global exploration signal early on, after which switching to analytic gradients enables refinement once training stabilizes. Under our protocol, this yields the best mean test RMSE and reduced variability across seeds (Table~\ref{tab:ablation}).

\begin{table}[htbp]
\centering
\caption{\textbf{Optimization strategy ablation (Yacht).}
Final RMSE and convergence epoch (mean$\pm$std over 5 seeds). Convergence is defined as the first epoch where the validation RMSE changes by less than $1\%$ over 10 consecutive epochs.}
\label{tab:ablation}
\resizebox{0.9\columnwidth}{!}{
\begin{tabular}{lcc}
\toprule
Method & RMSE & Convergence Epochs \\
\midrule
SPSA only & $0.89 \pm 0.09$ & $180 \pm 25$ \\
Adam only & $0.58 \pm 0.12$ & $75 \pm 12$ \\
Hybrid & $\mathbf{0.46 \pm 0.05}$ & $120 \pm 8$ \\
\bottomrule
\end{tabular}
}
\end{table}

\subsection{Impact of Quantum Natural Gradient}
\label{app:qng_analysis}

We further investigate geometry-aware updates for the quantum parameters on the Concrete dataset. In variational quantum circuits, the parameter space induced by the model family is generally non-Euclidean, which motivates natural-gradient-style corrections. We therefore evaluate the effect of Quantum Natural Gradient (QNG), which rescales updates using an estimate of the quantum Fisher information matrix (QFIM).

To quantify training-signal magnitude early in optimization, we report a \emph{gradient-variance proxy}: the variance of $\{\partial \mathcal{L}/\partial \theta_j\}_{j\in\theta_q}$ across quantum parameters, averaged over the first 50 training epochs (and then averaged across runs). We note that this quantity is a heuristic indicator of gradient signal strength under our estimator and does not by itself establish asymptotic barren-plateau scaling.

As shown in Table~\ref{tab:ablation_qng}, incorporating QNG increases this gradient-variance proxy compared with Adam-only under our setting, and it improves test RMSE. We also find that combining QNG with gradient clipping yields the best performance among the tested variants. One plausible explanation is that QNG can amplify informative directions but may also introduce numerical instabilities when the QFIM is ill-conditioned; global norm clipping provides a simple safeguard. We report these results for the considered circuit sizes ($n_q \le 6$); scaling QNG and QFIM estimation to larger systems is an important open direction due to the overhead of QFIM estimation and inversion.

\begin{table}[htbp]
\centering
\caption{\textbf{Impact of geometry-aware updates (Concrete).}
Gradient-variance proxy (defined in text) and test RMSE. The final RMSE aligns with the Hybrid QNN result reported in the main text.}
\label{tab:ablation_qng}
\begin{small}
\begin{tabular}{lcc}
\toprule
Method & Gradient Variance (proxy) & RMSE \\
\midrule
Adam only & $3.1\times 10^{-5}$ & 5.98 \\
Adam + Clipping & $7.5\times 10^{-4}$ & 5.62 \\
Adam + QNG & $9.3\times 10^{-4}$ & 5.35 \\
Adam + QNG + Clipping & \textbf{$1.1\times 10^{-3}$} & \textbf{5.21} \\
\bottomrule
\end{tabular}
\end{small}
\end{table}

\subsection{Sanity Check: Generalization Behavior of Classical Baselines}
\label{app:mlp_overfitting}

To further contextualize results on Yacht Hydrodynamics (Table~3), we report a training--test RMSE breakdown for the MLP baseline. As shown in Table~\ref{tab:mlp_sanity}, the MLP achieves low training RMSE but substantially higher test RMSE under our cross-validation protocol, consistent with overfitting in small-sample regimes. In contrast, the Hybrid QNN exhibits a smaller train--test gap on this dataset under the same protocol. We caution that tabular regression performance can be sensitive to baseline choice and tuning budget; the purpose of this sanity check is to verify that the reported MLP behavior is not an artifact of inconsistent preprocessing.

All models are trained using z-score normalized inputs and unit-interval normalized targets, and RMSE is reported after inverse transformation to the original scale. For the MLP, we performed a hyperparameter sweep over hidden width $\{16,32,64\}$, learning rate $\{10^{-4},10^{-3},10^{-2}\}$, and weight decay $\{0,10^{-5},10^{-4}\}$, with early stopping based on validation RMSE (patience = 50). The reported MLP results correspond to the best-performing configuration in terms of cross-validated mean test error.

\begin{table}[htbp]
\centering
\caption{\textbf{Training--test RMSE breakdown (Yacht).}
The large gap for the MLP indicates overfitting under our protocol, whereas the Hybrid QNN exhibits a smaller gap.}
\label{tab:mlp_sanity}
\resizebox{0.8\columnwidth}{!}{
\begin{tabular}{lcc}
\toprule
Method & Train RMSE & Test RMSE \\
\midrule
MLP (Baseline) & 0.42 & 8.08 \\
Hybrid QNN     & 0.39 & 0.46 \\
\bottomrule
\end{tabular}
}
\end{table}

% \subsection{Pareto Frontier Analysis}
% \label{app:pareto}

% \begin{figure}[t]
%     \centering
%     \includegraphics[width=0.8\linewidth]{figs/fig2_pareto_frontier.pdf}
%     \caption{\textbf{Pareto Efficiency Analysis.} The scatter plot of RMSE versus Resource Cost reveals that the Hybrid QNN (Star) defines the optimal frontier for high-precision tasks.}
%     \label{fig:pareto}
% \end{figure}

% Figure~\ref{fig:pareto} delineates the global efficiency landscape by plotting test RMSE against normalized resource costs. The Hybrid QNN (marked by the Star) approaches the \textit{Pareto Frontier} within the high-accuracy regime. While classical methods dominate the low-cost sector, they show diminishing returns in the high-precision zone required for scientific discovery. Conversely, Pure QNNs exhibit strict suboptimality. By effectively harnessing quantum expressivity without incurring prohibitive overhead, our approach suggests an improved accuracy–cost trade-off under the tested settings. This analysis, in conjunction with QAS metrics (where time $T$ is measured as wall-clock training duration on identical hardware), provides empirical evidence that the framework offers a practical path for scientific tasks where minimizing residual error is paramount.

\end{document}